\documentclass[letterpaper, 10 pt, conference]{IEEEtran/ieeeconf}  

\IEEEoverridecommandlockouts                              

\usepackage{cite}
\usepackage{amsmath,amssymb,amsfonts}
\usepackage{algorithmic}
\usepackage{algorithm}
\usepackage{graphicx}
\usepackage{tabularx}
\usepackage{textcomp}
\usepackage{xcolor}
\usepackage{stfloats}
\usepackage{booktabs} 
\usepackage{url}
\usepackage{verbatim}
\usepackage{nccmath}
\usepackage{mathtools}
\usepackage{array}
\usepackage[margin=1in]{geometry}

\renewcommand{\arraystretch}{1.1}

\newcolumntype{C}[1]{>{\centering\arraybackslash}p{#1\linewidth}}
\newcolumntype{L}[1]{>{\raggedright\arraybackslash}p{#1\linewidth}}
\newcommand{\bb}[1]{\boldsymbol{#1}}
\DeclareMathOperator*{\argmax}{arg\,max}
\DeclareMathOperator*{\argmin}{arg\,min}

\title{\LARGE \bf
Toward Global Intent Inference for Human Motion by Inverse Reinforcement Learning
}

\author{Sarmad Mehrdad$^{1}$, Maxime Sabbah$^{2}$, Vincent Bonnet$^{2,3}$, Ludovic Righetti$^{1,4}$
\thanks{${^1}$  Machines in Motion Laboratory, New York University, USA}
\thanks{${^2}$ LAAS-CNRS, Université Paul Sabatier, CNRS, Toulouse, France}
\thanks{${^3}$Image and Pervasive Access Laboratory (IPAL), CNRS-UMI, Singapore}
\thanks{${^4}$ Artificial and Natural Intelligence Toulouse Institute (ANITI), Toulouse}
}

\begin{document}

\maketitle
\thispagestyle{empty}
\pagestyle{empty}

\begin{abstract}
This paper investigates whether a single, unified cost function can explain and predict human reaching movements, in contrast with existing approaches that rely on subject- or posture-specific optimization criteria. Using the Minimal Observation Inverse Reinforcement Learning (MO-IRL) algorithm, together with a seven-dimensional set of candidate cost terms, we efficiently estimate time-varying cost weights for a standard planar reaching task. MO-IRL provides orders-of-magnitude faster convergence than bilevel formulations, while using only a fraction of the available data, enabling the practical exploration of time-varying cost structures.
Three levels of generality are evaluated—Subject-Dependent Posture-Dependent, Subject-Dependent Posture-Independent, and Subject-Independent Posture-Independent. Across all cases, time-varying weights substantially improve trajectory reconstruction, yielding an average 27\% reduction in RMSE compared to the baseline. The inferred costs consistently highlight a dominant role for joint-acceleration regulation, complemented by smaller contributions from torque-change smoothness.
Overall, a single subject- and posture-agnostic time-varying cost function is shown to predict human reaching trajectories with high accuracy, supporting the existence of a unified optimality principle governing this class of movements.

\end{abstract}

\section{Introduction}
In human–robot interaction and collaborative manipulation, robots benefit from inferring human intention early during motion. Even simple reaching movements can convey actionable intent, such as target selection, obstacle avoidance, or context-dependent preferences. If these cues can be extracted from partial trajectories, a robot can proactively yield, assist, or generate safe motions, rather than reacting to the observed movement.

Despite kinematic redundancy, human motion exhibits robust invariants, suggesting that the nervous system resolves redundancy through consistent organizational principles often described by optimal cost trade-offs \cite{maas2013biomechanical, alexander1984gaits}. A natural way to capture such intent is therefore to model human motion within an optimal control framework \cite{todorov2002optimal, todorov2004optimality}. While humans are not strictly optimal, this formulation provides a compact and physically grounded language to describe motion regularities through interpretable cost features. Identifying these features enables both explanation of human motion strategies and generation of similar movements in new contexts. In this setting, Inverse Optimal Control (IOC) \cite{mombaur2010human} and Inverse Reinforcement Learning (IRL) \cite{arora2021survey} are the main tools used to infer task principles from demonstrations.

Although the existence of optimality principles is widely recognized \cite{pandy1990optimal, xiang2010physics, sylla2014human, bevcanovic2022gait}, how costs adapt within or across tasks remains unclear. For example, humans often slow down near the target to improve accuracy, revealing a trade-off between task-related and body-related costs that many models fail to capture \cite{zelik2012}. Moreover, most approaches rely on a single cost function per task \cite{Lin2021}, which can lead to large prediction errors. Sylla et al. \cite{sylla2014}, using a computationally expensive bi-level approach, reported average joint errors of about 7deg and values exceeding 30deg even for simple reaching tasks. Quantitative evaluations are also frequently missing or rely on tailored metrics \cite{berret2011}, raising concerns about predictive validity. While time-varying cost formulations have been proposed \cite{lin2016human, jin2019inverse, sabbah2025minimal}, they introduce many parameters and are typically individualized, limiting their ability to capture general principles of human motion.

A major limitation in cost inference is the difficulty of applying standard IOC and IRL methods directly to human data. IOC formulations rely on repeated solutions of nested optimal control problems, leading to prohibitive computational cost and sensitivity to local minima in high-dimensional settings. Alternatives based on Karush–Kuhn–Tucker residuals \cite{englert2017, bevcanovic2022gait} avoid nested optimization but remain highly sensitive to measurement noise and modeling errors \cite{colombel2022, bevcanovic2022}, which are inherent to human motion data.

IRL provides a probabilistic alternative with fewer nested optimizations, but its performance depends strongly on the quality of the trajectory-space approximation. Existing methods rely on local sampling \cite{kalakrishnan2011, kalakrishnan2013} or importance sampling \cite{finn2016}, producing trajectories that remain close to demonstrations while increasing computational cost and limiting exploration.

Recently, Minimal Observation Inverse Reinforcement Learning (MO-IRL) \cite{mehrdad2025} has been proposed to address these issues. MO-IRL combines a probabilistic IRL formulation with constrained optimal control, automatically weighting sampled trajectories based on their estimated optimality. Even with few observations, it achieves faster convergence and more informative weight updates.

In this paper, we investigate for the first time whether time-varying cost weights can recover a single unified cost function explaining human reaching motions across subjects and initial postures. Addressing this question requires an IRL method that remains efficient as the number of parameters increases. To this end, we extend MO-IRL to learn time-varying cost weights and propose an estimation procedure that jointly exploits position and velocity information from demonstrations. We validate our approach on a widely used human motor control dataset \cite{berret2011}.

\section{Methods}
\subsection{Experimental protocol and mechanical model}
The human motion data used in this study were provided by Berret et al. \cite{berret2011} and have since become a reference dataset in the literature. They consist of 3D motion-capture marker positions at the shoulder, elbow, and wrist, recorded from 15 right-handed naïve subjects.

\begin{figure}
    \centering
    \includegraphics[width=0.9\linewidth]{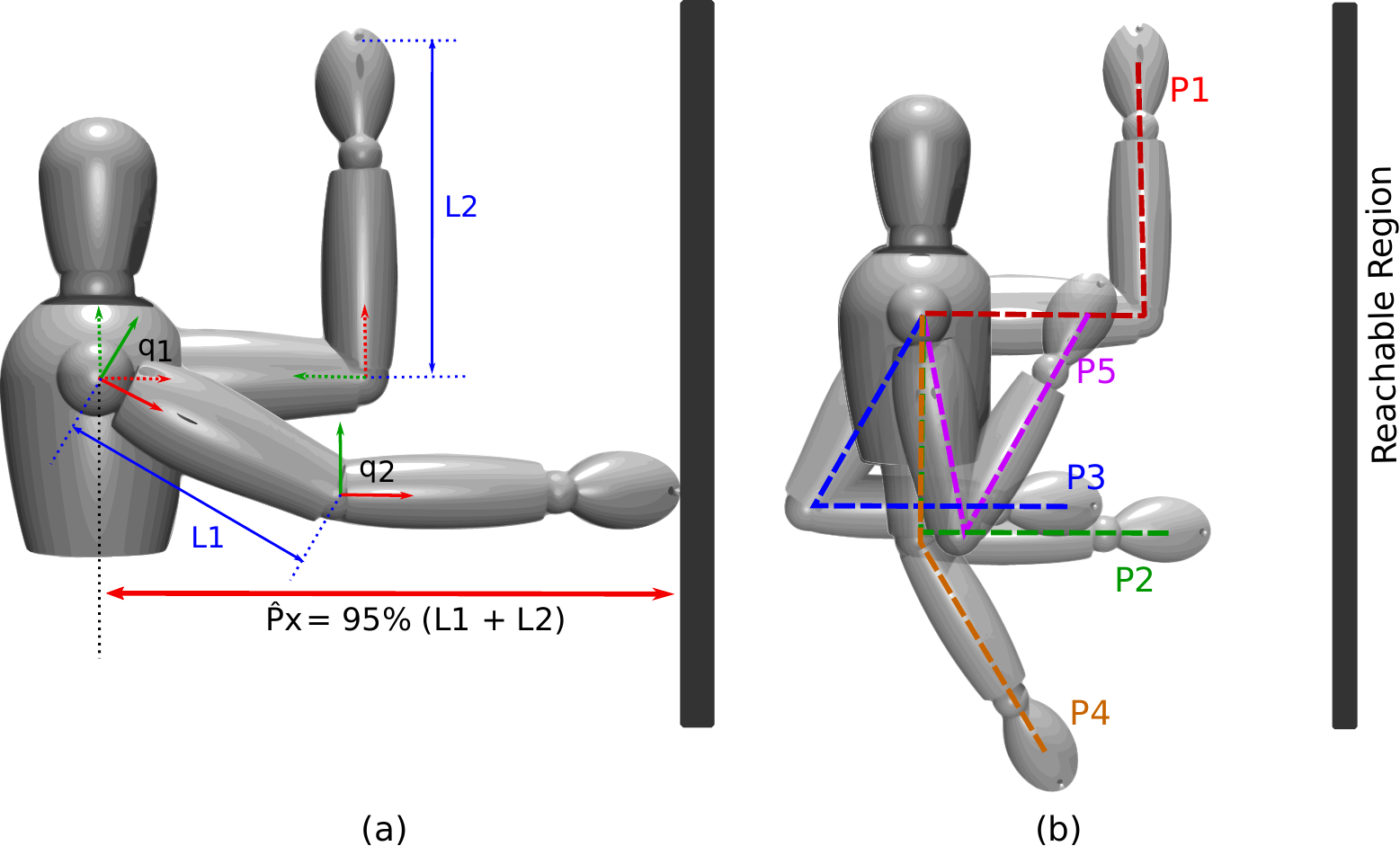}
    \caption{(a) Biomechanical model definition, showing the beginning and the end of the pointing task. (b) Five different initial postures for the pointing task \cite{berret2011}.}
    \label{fig:biomec_model}
\end{figure}

Subjects performed the pointing task depicted in Fig. \ref{fig:biomec_model}.a. While seated, participants were instructed to perform a series of pointing movements toward a vertical target bar positioned in front of the participant. Only shoulder and elbow flexion/extension were permitted during the task.  The shoulder-to-bar horizontal distance was set to 85\% of the participant’s total arm length ($L = L_1 + L_2$, with $L_1$ and $L_2$ representing upper arm and forearm lengths, respectively; Fig. \ref{fig:biomec_model}.a). Five initial arm postures, labeled P1 through P5 and shown in Fig. \ref{fig:biomec_model}.b, were defined. Each individual performed 20 trials for each posture.

A planar biomechanical model (Fig.\ref{fig:biomec_model}.a) was developed to represent flexion/extension movements at the shoulder ($q_1$) and elbow ($q_2$) joints. The model’s base frame was located at the shoulder joint. Inertial parameters were calculated using anthropometric tables \cite{Dumas2007}.

\subsection{Optimal control problem}
In the context of pointing or reaching movements, inspired by the literature in human motor control \cite{flash1985, nishii2002, berret2008, biess2007, ben2008, uno1989,nakano1999, atkeson1985, nelson1983}, we propose a set of $N_\Phi = 7$ candidate cost functions, as detailed in Table \ref{Table:cost_functions}. In this table, $\bb{q}$, $\bb{v}$, and $\bb{a}$ are human arm joint values, velocities, and acceleration respectively. $M$ is the mass matrix of the human arm linkage, $\bb{V}$ is the cartesian velocity vector, and $\bb{\tau}$ is the joint torque vector. Although the task may appear elementary, we posit that individuals do not adhere to a single cost function throughout the entire movement. To account for such time-varying motor strategies, each recorded trajectory of duration $T$ was segmented into $N_w$ equal time windows, each comprising $N_s$ samples ($T = N_w \times N_s$). This segmentation was chosen based on consistent inflection points observed in the majority of trajectories.
To model the temporal evolution of movement strategies, we introduced a weight matrix $\bb{w} \in \bb{R}^{N_w \times N_\Phi}$, which allows distinct cost function contributions across different movement phases. The full trajectory $\bb{x}_t \in \bb{R}^{2N_{d}}, \hspace{2mm} t = 0, 1, \dots, T$ was defined as the concatenation of state vectors $\bb{x}_t = (\bb{q}_t, \bb{v}_t)$ over time, with $N_d = 2$ being the degrees of freedom for the human arm in our case. Accordingly, $\bb{u}_t \in \bb{R}^{N_{d}} , \hspace{2mm} t = 0, 1, \dots, T-1$ was defined as the control torque input to the human model joints for each timestamp. The associated Direct Optimal Control (DOC) problem was then formulated to reconstruct the observed human motion over this multi-phase framework.

\begin{equation}
\begin{aligned}
\bb{x}^*, \bb{u}^* = \ & \argmin_{\bb{u}} \quad \sum_{s=1}^{N_w} \bb{w}_s ^T \bb{\Phi}_s \\
\text{s.t.} & \quad \bb{\Phi}_s = \sum_{t=(s-1)N_s}^{sN_s -1} \Phi(\bb{x}_t, \bb{u}_t) \\
            & \quad \bb{x}_{t+1} = f(\bb{x}_t, \bb{u}_t) \hspace{2mm} \scriptsize{\texttt{Dynamics}} \\
            & \quad FK(\bb{x}_T)_X = \hat{P}_{X}  \hspace{2mm} \scriptsize{\texttt{Forward Kinematics}} \\
            & \quad \bb{q}^-\leq \bb{q}_t \leq \bb{q}^+\\
            & \quad \bb{q}_{t}=\hat{\bb{q_0}}, \hspace{2mm} t = 0\\
            & \quad |\bb{v}_t|\leq\bb{v}^+\\
\end{aligned}
\label{eq:doc}
\end{equation}

\noindent where $\bb{x}_t$ and $\bb{u}_t$ are the state and control at discrete time $t$; $f$ is the time-discretized dynamics; $\hat{P}_X$ is the horizontal goal position (Fig. \ref{fig:biomec_model}.a) obtained from forward kinematics. $\hat{\bb{q_0}}$ is the initial human joint configuration; $\bb{q}^-$, $\bb{q}^+$ are the lower and upper joint boundaries,  respectively; $\bb{v}^+$ is the maximal joint velocity. 

Modeling of the human body was done using Pinocchio \cite{pin}, and the Croccoddyl framework \cite{crocoddyl} and the MiM\_Solver nonlinear optimal control solver \cite{sqp} were used to define and solve the constrained DOC in Eq.\eqref{eq:doc}. 


\begin{table}[!t]
\centering
\fontsize{7}{11}\selectfont
\caption{(Discretized) Biomechanical cost functions}
\begin{tabular}{clll} \toprule
\textbf{Label} &\textbf{Name} & \textbf{Equation} & \textbf{Reference} \\ \midrule
$\Phi_1$    & Cartesian velocity  (CV)   &  $ \sum_{t=0}^{T} \bb{V}_t^T\bb{V}_t \,\Delta t $ & \cite{flash1985}     \\
$\Phi_2$    & Energy (Eng)    &  $ \sum_{t=0}^{T} \left|\bb{v}_t^T \mathbf{\bb{\tau}}_t\right| \Delta t $ & \cite{nishii2002, berret2008}     \\
$\Phi_3$    & Geodesic (Geo)    &  $ \sum_{t=0}^{T} \bb{v}_t^TM(\bb{q}_t) \bb{v}_t \Delta t $ & \cite{biess2007}     \\
$\Phi_4$    & Joint acceleration (JA)    &  $ \sum_{t=0}^{T} \bb{a}_t^T\bb{a}_t \Delta t $ & \cite{ben2008}     \\
$\Phi_5$    & Joint torque change (JTC)    &  $ \sum_{t=0}^{T} \dot{\bb{\tau}}_t^T\dot{\bb{\tau}}_t \Delta t$ & \cite{uno1989,nakano1999}    \\
$\Phi_6$    & Joint velocity (JV)    &  $ \sum_{t=0}^{T} \bb{v}_t^T\bb{v}_t  \Delta t $ & \cite{atkeson1985}     \\
$\Phi_7$    & Joint torque (Tau)    & $ \sum_{t=0}^{T} \bb{\tau}_t^T\bb{\tau}_t  \Delta t $ & \cite{nelson1983}     \\ \bottomrule
\end{tabular}
\label{Table:cost_functions}
\end{table} 

\subsection{Minimum Observation Inverse Reinforcement Learning (MO-IRL)}

We extend the existing MO-IRL algorithm \cite{mehrdad2025} to varying cost weights across time windows, as defined in Eq.~\eqref{eq:doc}. IRL algorithms generally maximize the probability of the optimal demonstrations (i.e. the human trajectory)

\begin{align}
    \label{maxent}
    &\bb{w}^* = \argmax_{\bb{w}} P(\bb{\xi}^*|\bb{w} , \Bar{\bb{\xi}}) \\
    \text{where} \hspace{5mm} &P(\bb{\xi}^*|\bb{w} , \Bar{\bb{\xi}}) = \frac{e^{-\bb{w}^T\bb{\Phi}^*}}{\sum_{\bb{\xi}_i \in \Bar{\bb{\xi}}} e^{-\bb{w}^T\bb{\Phi}_i}} \nonumber \\
    &\bb{w} \geq 0 \nonumber
\end{align}

\noindent
in which $\bb{\xi}$ refers to trajectories consisting of the state and control vector pairs ($\bb{\xi}_i = \left\langle \bb{x}^i, \bb{u}^i \right\rangle$), and $\Bar{\bb{\xi}}$ is the set of $K$ observed trajectories. For clarity, we write $\bb{w}$ the concatenated weight vector for the cost features and $\bb{\Phi}$ the concatenated feature vector. In the following, we refer to the feature costs for the $i$\textsuperscript{th} trajectory $\bb{\xi}_i$ as $\bb{\Phi}_i$.

MO-IRL solves Eq. \eqref{maxent} by iteratively improving $\bb{w}$ rather than optimizing it in one shot. Considering an update of cost weights at each iteration $n+1$ in the form $\bb{w}_{n+1} = \bb{w}_n + \Delta\bb{w}_n$, the original probability distribution can be rewritten to instead find the best $\Delta\bb{w}_n$:

\begin{align}
    \label{'delta_w'}
    \Delta\bb{w}_n &= \argmin_{\Delta\bb{w}} -\log \frac{1}{1 + \sum_{\bb{\xi}_i \in \Bar{\bb{\xi}}} \gamma_i e^{-\Delta\bb{w}^T(\bb{\Phi}_i - \bb{\Phi}^*)}} \nonumber\\
    \text{s.t.}  \hspace{3mm} &\Delta\bb{w} > -\bb{w}_n \\
    &\gamma_i = e^{-\bb{w}_n^T(\bb{\Phi}_i - \bb{\Phi}^*)} \nonumber
\end{align}

\noindent
In this case, sampled trajectories are automatically scaled depending on their cost in the previous iteration. MO-IRL solves Eq. \eqref{'delta_w'} and then seeks to find an update of the form $\bb{w}_{t+1} = \bb{w}_t + \alpha\Delta \bb{w}$ where $\alpha$ is selected using a merit function (similar to a line search procedure). Starting with $\alpha = 1$, the algorithm checks if the resulting trajectory is closer to the optimal demonstration by evaluating the merit function. If the merit function value has not been decreased with the added change to the weight, MO-IRL scales down $\alpha$ by factor of $0.25$, and tries again for a maximum of 10 trials. If by the 10\textsuperscript{th} trial there was no improvement, the algorithm stops. Otherwise, the accepted trajectory is added to the observed trajectory set $\Bar{\bb{\xi}}$, and MO-IRL moves on to the next iteration.

In the literature, algorithms to learn human motion trajectories usually minimize the gap between the estimated and the optimal trajectories in joint space without considering velocities. In this study, however, we propose to minimize the gap in both joint position and joint velocity concurrently. Therefore, the estimation improvement was based on the full state vector $\bb{x} = [q_1, q_2, v_1, v_2]$ with   $m(\bb{x}) = \frac{1}{T}||\bb{x}^* - \bb{x}||^2_2$ as merit function. We extend the MO-IRL algorithm to work with multiple weight sections as follows:
\begin{align}
    \label{'moirl'}
    \small
    &\Delta \bb{w}_n =  \argmin_{\Delta \bb{w}} \nonumber \\
    &\sum_{d = 1}^D \bigg( -\log \frac{1}{1 + \sum_{\bb{\xi}_i \in \Bar{\bb{\xi}}} \gamma_i e^{-C(\bb{\xi}_i, \Delta\bb{w})}}\bigg) + \frac{\beta}{2}||\Delta \bb{w}||_2^2 \nonumber \\
    &\text{s.t.}  \hspace{3mm} \Delta\bb{ w} > -\bb{ w}_n  \\
    &\hspace{7mm} \gamma_i = e^{-C(\bb{\xi}_i, \bb{w}_n)} \nonumber \\
    &\hspace{7mm} C(\bb{\xi}_i, \bb{w}) = \sum_{s = 1}^{N_ w} \bb{w}_{s}^T(\Phi_{si} - \Phi_{sd}^*) \nonumber
    \normalsize
\end{align}
%
where $D$ is the number of provided optimal demonstrations. We also use a small $L2$ regularizer ($\beta = 10^{-9}$) for the optimization to prevent high changes in weights and overfitting. When learning from multiple demonstrations, the merit function for  step acceptance is changed to $m(\bb{x}) = \frac{1}{D}\sum_{d = 1}^D(\frac{1}{T_d}||\bb{x}_d^* - \bb{x}||^2_2)$. In case the input demonstrations are from various initial postures, MO-IRL will generate one trajectory per initial posture to evaluate the merit function value against human demonstrations starting from the same posture. Eventually, all generated trajectories from different initial postures will be added to $\Bar{\bb{\xi}}$ when the step gets accepted.


\subsection{Learning and Validation}

In this study, the proposed approach was validated across three cases  ranging from subject and posture specific cost functions generation to a generic one. We consider $N_w = \Big\lfloor \tfrac{1}{2} T_d \Big\rfloor$ weight sections for each demonstration with the length $T_d$, to have maximum number of weight sections allowable while still enabling MO–IRL to identify meaningful time variations in the cost structure. Note that if $N_w > \Big\lfloor \tfrac{1}{2} T_d \Big\rfloor$, at least one section will have length of $1$, which will be zeroed out.

We evaluate the quality of the weights estimated by MO-IRL by computing the RMSE between the resulting predicted trajectory (computed by solving the optimal control problem) and the human demonstration. For each case, we consider a cost function to be better if it would regenerate human motions that closely matched the recorded demonstrations.


\subsubsection{Subject-Dependent Posture-Dependent (SDPD)}

Similarly to Berret et al. \cite{berret2011}, SDPD analysis aims to estimate the optimal cost function specified for each subject for each initial posture. The input non-optimal set in this case consists of 3 random demonstrations for each subject and each posture. Overall, there are $15 \text{(subject)} \times 5 \text{(postures)} = 75$ estimated cost functions.

\subsubsection{Subject-Dependent Posture-Independent (SDPI)}

The SDPI case evaluates if optimal weights can be estimated irrespectively of the initial arm posture. In other words, we aim to find subject specific weights for the reaching task. We use a set of 5 demonstrations as input for MO-IRL, consisting of one random demonstration each posture. Overall, there will be $15 \text{(subject)}$ estimated cost functions resulting from this test case.

\subsubsection{Subject-Independent Posture-Independent (SIPI)}

The SIPI case aims to estimate cost function irrespective of the subject and initial posture. In this case, the input demonstration set consists of 5 random demonstration from each subject (one per initial posture), totaling 75. This case will produce one set of estimated cost weights.

\section{Results}

\begin{figure*}[t]
   \centering
   \includegraphics[width=0.85\linewidth]{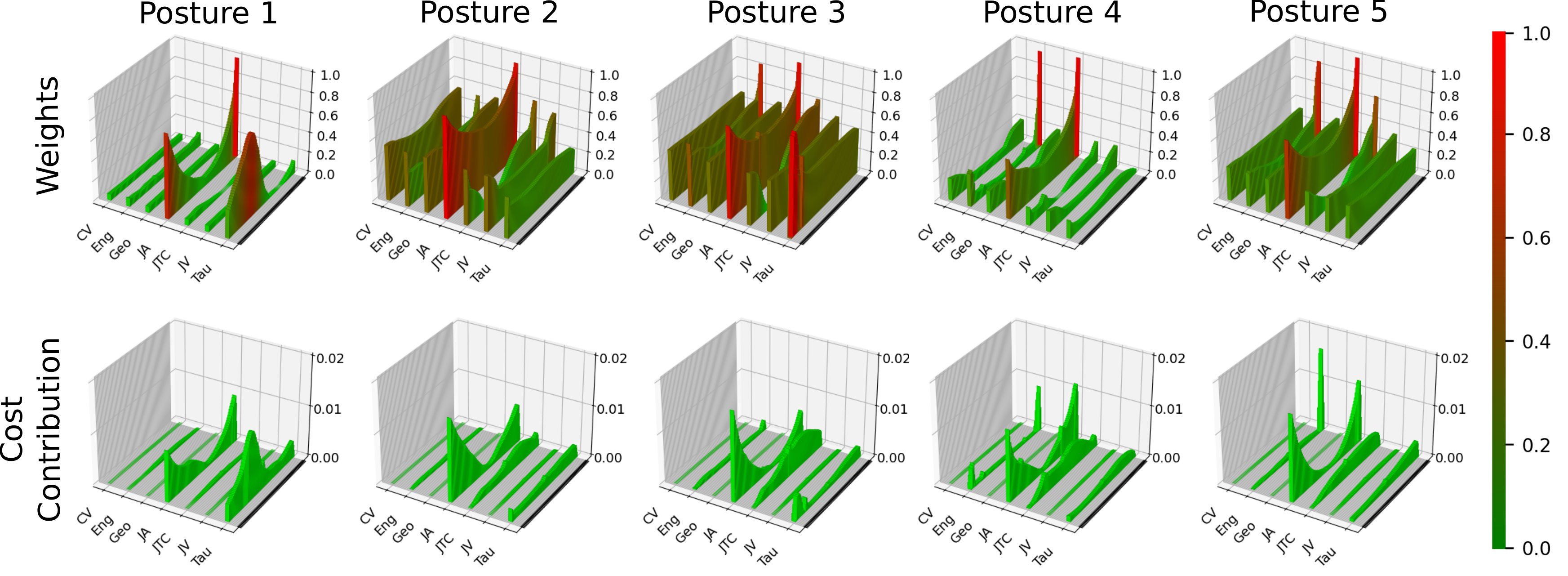}
   \caption{
   Averaged estimated weights, and cost contribution, across all subjects for initial postures 1-5, produced through SDPD analysis. Plots show a prominent importance of the Joint Acceleration feature in the cost function ($\Phi_4$).}
   \label{fig:SDPD_avg}
\end{figure*}

\begin{figure*}[t]
   \centering
   \includegraphics[width=0.85\linewidth]{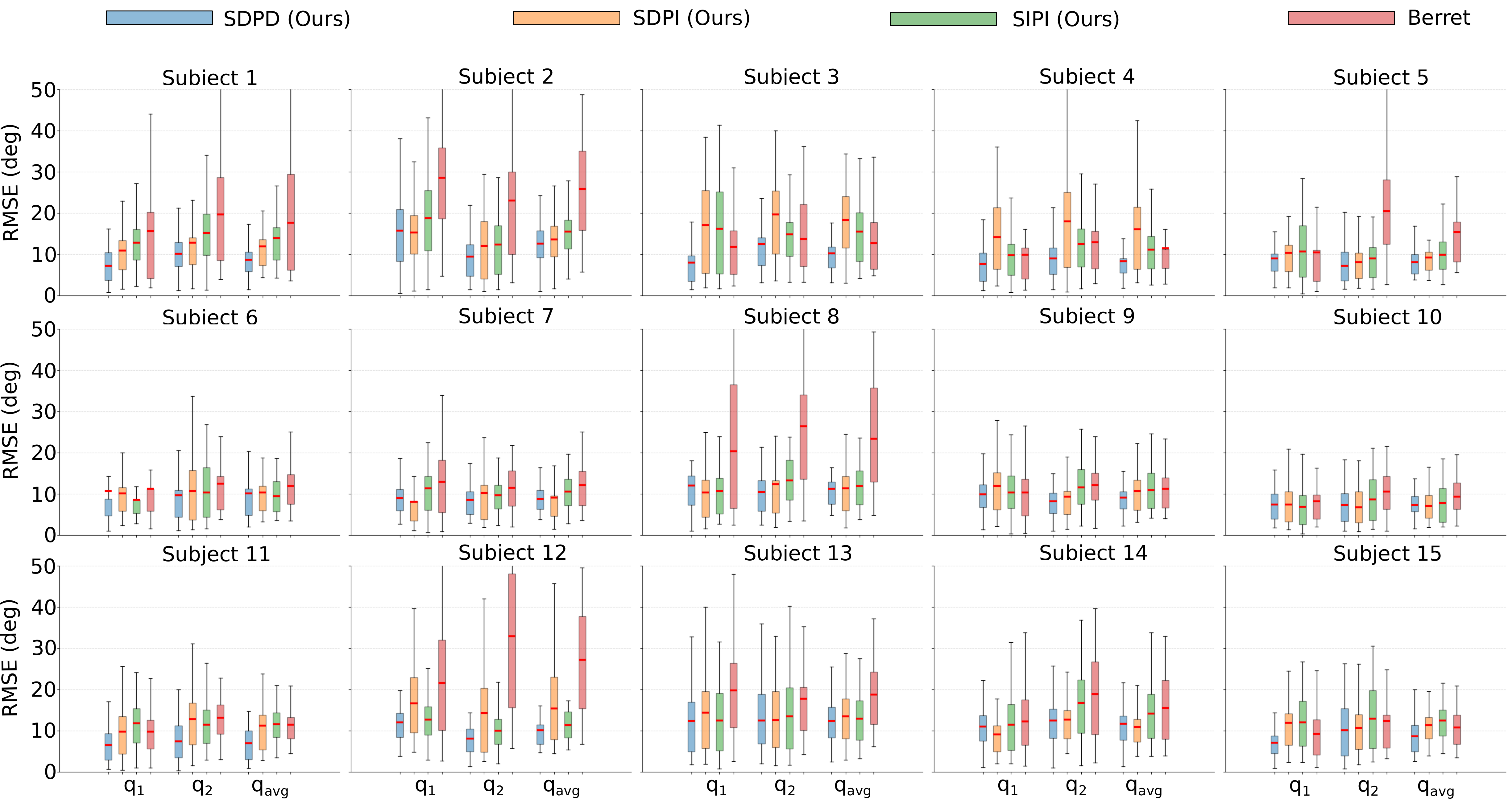}
   \caption{RMSE values boxplot comparison of all investigated cases against the baseline, averaged across all subjects. Red line in each boxplot indicates the mean value.}
   \label{fig:subjects}
\end{figure*}

\begin{figure*}[t]
   \centering
   \includegraphics[width=0.85\linewidth]{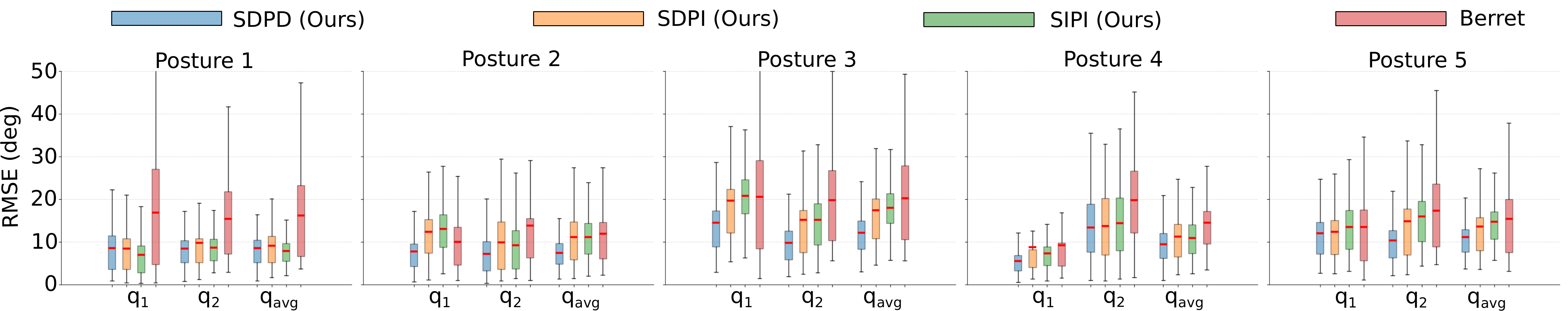}
   \caption{RMSE values (deg) boxplot comparison of all methods against the baseline, averaged across all initial postures. Red line in each boxplot indicates the mean value.} 
 
   \label{fig:postures}
\end{figure*}

Fig.~\ref{fig:SDPD_avg} presents the average estimated weights for all subjects and postures in the SDPD condition. A consistent pattern emerges across all initial postures, with the acceleration cost term $\Phi_4$ dominating the overall contribution. The estimated acceleration weight is particularly high near the beginning and end of the reach, while remaining lower during mid–movement. The resulting predicted joint trajectories exhibit and average RMSE values of $9.59 \pm 5.20$ deg, as evident in Table \ref{tab:SDPD_num}. This is significantly lower than values reported in the literature using a single weight set \cite{berret2011, sylla2014, Lin2021}. Using the  fixed weights, obtained from a bi-level approach, for the same pointing data, Berret et al. obtained an average RMSE of $15.44 \pm 10.57$ deg. Figs.~\ref{fig:subjects} and \ref{fig:postures} further demonstrate that time varying weights systematically improve prediction accuracy across subjects and postures.

\begin{table}[h]
\centering
\scriptsize
\setlength{\tabcolsep}{5pt}
\renewcommand{\arraystretch}{1.1}
\begin{tabular}{|l|ccc|ccc|}
\hline
& \multicolumn{3}{c|}{MO-IRL SDPD (Ours)} & \multicolumn{3}{c|}{Baseline} \\
Initial Posture &  $q_1$ & $q_2$ & Avg &  $q_1$ & $q_2$ & Avg \\
\hline
P1 & 8.56 & 8.47 & \textbf{8.51} & 16.92 & 15.40 & 16.16 \\
P2 & 7.77 & 7.17 & \textbf{7.47} & 10.06 & 13.84 & 11.95 \\
P3 & 14.46 & 9.76 & \textbf{12.11} & 20.60 & 19.83 & 20.21 \\
P4 & 5.52 & 13.33 & \textbf{9.43} & 9.28 & 19.84 & 14.56 \\
P5 & 11.99 & 10.36 & \textbf{11.17} & 13.54 & 17.35 & 15.45 \\
All & 9.50 & 9.68 & \textbf{9.59} & 13.89 & 16.99 & 15.44 \\
\hline
\end{tabular}
\caption{RMSE results of SDPD weights for reaching task generated by MO-IRL, in comparison with the baseline.}
\label{tab:SDPD_num}
\end{table}

In the SDPI condition, a single time–varying weight profile is learned per subject across all postures. The resulting weights again show an obvious  rise and fall in $\Phi_4$, confirming the central role of acceleration regularization independently of the initial configuration. RMSE values for each posture are summarized in Table~\ref{tab:SDPI_num}, showing that SDPI outperforms the baseline in almost all joints and postures. This is an other clue supporting that  allowing time–varying weights provides a more flexible representation of human motor patterns than a static cost structure.

\begin{table}[h]
\centering
\scriptsize
\setlength{\tabcolsep}{5pt}
\renewcommand{\arraystretch}{1.1}
\begin{tabular}{|l|ccc|ccc|}
\hline
& \multicolumn{3}{c|}{MO-IRL SDPI (Ours)} & \multicolumn{3}{c|}{Baseline} \\
Initial Posture &  $q_1$ & $q_2$ & Avg &  $q_1$ & $q_2$ & Avg \\
\hline
P1 & 8.39 & 9.80 & \textbf{9.10} & 16.92 & 15.40 & 16.16 \\
P2 & 12.40 & 9.94 & \textbf{11.17} & 10.06 & 13.84 & 11.95 \\
P3 & 19.71 & 15.17 & \textbf{17.44} & 20.60 & 19.83 & 20.21 \\
P4 & 8.75 & 13.76 & \textbf{11.26} & 9.28 & 19.84 & 14.56 \\
P5 & 12.32 & 14.83 & \textbf{13.57} & 13.54 & 17.35 & 15.45 \\
All & 11.99 & 10.36 & \textbf{11.17} & 13.89 & 16.99 & 15.44 \\
\hline
\end{tabular}
\caption{RMSE results of SDPI weights for reaching task generated by MO-IRL, in comparison with the baseline.}
\label{tab:SDPI_num}
\end{table}

\begin{figure}
   \centering
   \includegraphics[width=0.8\linewidth]{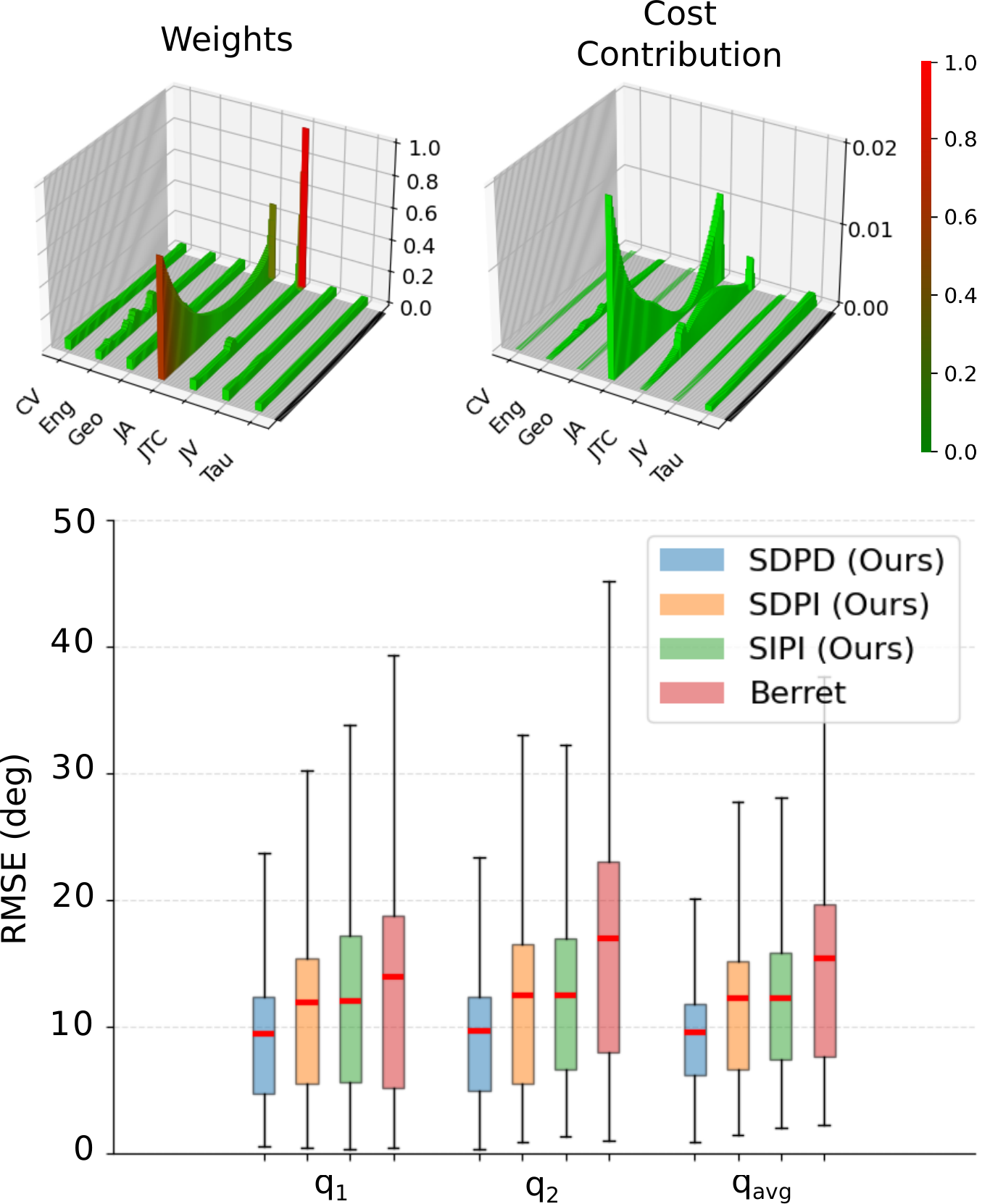}
   \caption{Weights and cost contributions of general cost weights proposed for the reaching task in SIPI case. The RMSE comparison boxplots between all existing demonstrations and predictions are presented in the figure at the bottom.}
   \label{fig:SIPI}
\end{figure}

\begin{figure}[t]
   \centering
   \includegraphics[width=0.6\linewidth]{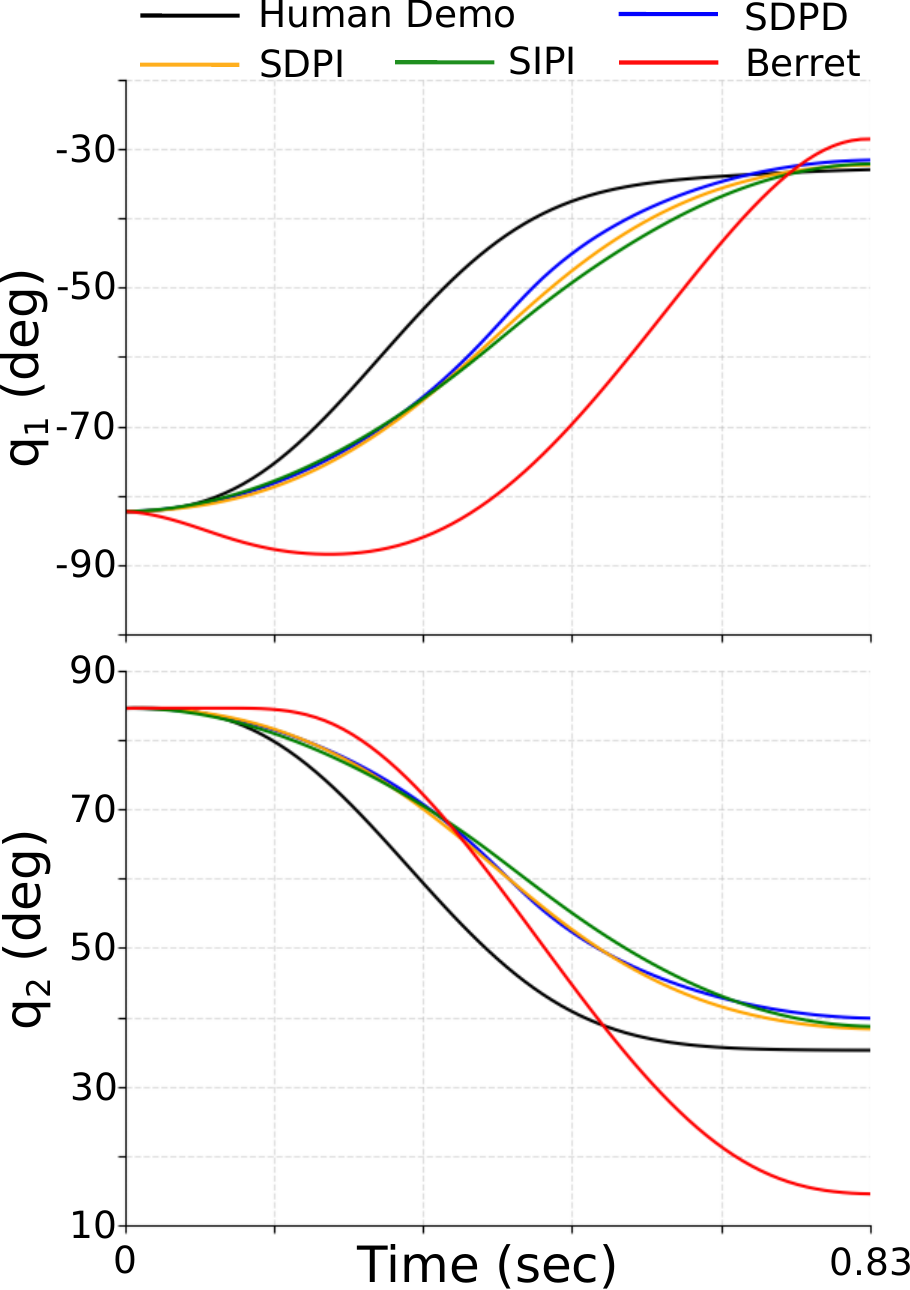}
   \caption{Comparison of SDPD, SDPI, SIPI, and baseline predictions of the subject's joint values, against the human demonstration.} 
 
   \label{fig:reaching}
\end{figure}

The SIPI condition investigates whether a single cost for the reaching task, independent of both subjects and initial postures can be recovered. The estimated weight profile and cost contributions are shown in Fig.~\ref{fig:SIPI}. The same temporal pattern for $\Phi_4$ is present at the population level, suggesting a strong commonality in how humans regulate acceleration. 

\begin{table}[h]
\centering
\scriptsize
\setlength{\tabcolsep}{5pt}
\renewcommand{\arraystretch}{1.1}
\begin{tabular}{|l|ccc|ccc|}
\hline
& \multicolumn{3}{c|}{MO-IRL SIPI (Ours)} & \multicolumn{3}{c|}{Baseline} \\
Initial Posture &  $q_1$ & $q_2$ & Avg &  $q_1$ & $q_2$ & Avg \\
\hline
P1 & 6.98 & 8.69 & \textbf{7.83} & 16.92 & 15.40 & 16.16 \\
P2 & 13.09 & 9.26 & \textbf{11.17} & 10.06 & 13.84 & 11.95 \\
P3 & 20.85 & 15.23 & \textbf{18.04} & 20.60 & 19.83 & 20.21 \\
P4 & 7.37 & 14.36 & \textbf{10.86} & 9.28 & 19.84 & 14.56 \\
P5 & 13.55 & 16.00 & \textbf{14.77} & 13.54 & 17.35 & 15.45 \\
All & 11.99 & 10.36 & \textbf{11.17} & 13.89 & 16.99 & 15.44 \\
\hline
\end{tabular}
\caption{RMSE results of SIPI weights for reaching task generated by MO-IRL, in comparison with the baseline.}
\label{tab:SIPI_num}
\end{table}

The SIPI results in Table~\ref{tab:SIPI_num} provide several important insights into the possible existence of a unified cost function for explaining human reaching motion across subjects and initial postures. First, the MO-IRL weights consistently outperform the baseline across all initial postures, despite being learned without access to subject- or posture-specific structure. The average RMSE is reduced by 27.65\% when compare to the baseline, which consider a subject-dependent cost function. This shows that the temporal profile of the inferred cost function captures fundamental aspects of human motor behavior that generalize across individuals and kinematic configurations.

Second, the reductions in RMSE are particularly substantial for postures P1 and P4. These postures involve relatively extended initial arm configurations, where small variations in joint torques likely lead to more predictable and less noisy task-space motions. In these cases, the general time-varying cost function appears sufficient to reproduce the smooth, bell-shaped trajectories characteristic of human reaches \cite{flash1985}, suggesting that the dominant control principles governing these movements are shared across subjects.

Posture P1 illustrates this point most strongly: the baseline method yields an average error of $16.16$~deg, while the SIPI cost function reduces this to $7.83$~deg. This large improvement indicates that static cost weights, such as those proposed in \cite{berret2011}, fail to capture the temporal modulation of control effort and smoothness that humans naturally apply.

Third, posture P3 yields the highest errors for both the baseline and SIPI conditions, reflecting the inherent difficulty of this configuration. Its more flexed-elbow posture requires larger joint excursions for small task-space motions, amplifying joint-space variability and the effects of signal-dependent motor noise \cite{harris1998}. Despite this, the SIPI weights still offer a clear improvement over the baseline, indicating that the general cost function captures consistent behavioral structure even under unfavorable dynamics.

Beyond the dominant influence of joint acceleration, the estimated weights also reveal a meaningful contribution of the joint torque change term ($\Phi_{5}$), visible in Fig.~\ref{fig:SIPI} through elevated mid-movement values. 

Fig. \ref{fig:reaching} shows the comparison of predictions using SDPD, SDPI, and SIPI methods against human data, for an example demonstration conducted by Subject 1 initiated from Posture 2. This example shows the effect of the gradual decline of specificity from SDPD to SDPI and ultimately SIPI, as the predicted trajectory deviates from the human demonstration further. However, the results in Fig. \ref{fig:reaching} show a significant improvement in comparison with the baseline in the joint space.

\section{Discussion}

Across all three evaluation cases, a consistent and robust pattern emerges in the inferred cost structure for human reaching. The joint acceleration term $\Phi_{4}$ is the dominant contributor, with a characteristic increase at movement onset and termination. This behavior aligns with classical motor–control findings: penalizing large accelerations avoids impulsive motor commands and reduces signal–dependent noise \cite{harris1998}, while the terminal increase reflects the need for precise endpoint stabilization, consistent with submovement corrections and bell–shaped velocity profiles \cite{flash1985, novak2002}.

A second key result is the non-negligible contribution of the joint torque change term $\Phi_{5}$ during the mid-reach phase. This observation complements and extends the acceleration-based explanation. The minimum–torque–change model of Uno et al.~\cite{uno1989} established that penalizing rapid variations in torques generates smooth and human-like multi-joint trajectories. More recent studies have further shown that stabilizing torque fluctuations helps reduce motor noise, improves robustness to interaction torques, and limits neuromuscular effort \cite{harris1998, dideriksen2021}. For our reaching scenario the joint torque change appearance at the middle of the motion is consistent with the need to maintain stable coordination while the segments are accelerating most rapidly, a period where motor noise and interaction torques are largest. The combined presence of $\Phi_{4}$ and $\Phi_{5}$ therefore suggests that human reaching optimizes both kinematic smoothness (acceleration regulation) and actuation smoothness (torque-change regulation), revealing a multi-criteria optimality principle that adapts over the course of the movement.

The limited influence of energy-related terms stands in contrast to the interpretation proposed in \cite{berret2011}. Several factors may explain this discrepancy. First, the pivot-based normalization used in \cite{berret2011} imposes strong constraints on the relative magnitudes of the cost weights, whereas the present formulation allows them to evolve freely over time. Second, the very large RMSE values reported when using Berret's weights (Table III) suggest that the attribution of a dominant role to joint energy may simply not be reliable. When prediction errors are so high, many different combinations of cost terms can yield equally poor fits to the data, making it difficult to draw firm conclusions about the underlying optimality principle. In this sense, the explanation based on joint energy may simply not be identifiable within the sensitivity limits of the DOC problem.

An other methodological difference from previous IOC and IRL studies concerns the use of both joint positions and joint velocities in the learning process. Most approaches in the literature rely exclusively on joint positions when estimating cost weights, either because velocities exhibit higher trial-to-trial variability or because position-only models simplify the inverse problem \cite{sylla2014}. However, excluding velocity information can obscure important aspects of the movement dynamics. Human reaching is characterized by highly stereotyped velocity profiles, and these profiles carry essential information about smoothness, timing, and coordination. Incorporating velocities therefore provides richer constraints on the underlying cost structure and reduces the ambiguity inherent in position-only formulations, where many weight combinations can reproduce similar end-point trajectories. In the present work, we believe that, the inclusion of velocity data contributes to the clear emergence of both acceleration and torque-change related terms, and likely improves the identifiability of time-varying weights. 

Taken together, these findings support the interpretation that the central nervous system relies on time-varying adjustments of a small set of key control principles, rather than on fixed task- or subject-specific cost parameters. The consistency of the inferred weights across heterogeneous initial postures highlights MO-IRL's ability to uncover this shared temporal structure.

Finally, the SIPI results demonstrate that a single time-varying cost function can reconstruct human trajectories across all subjects and postures with markedly higher accuracy than classical single-weight models. Achieving the performance reported in Table~\ref{tab:SIPI_num} with a fully general cost function is noteworthy, and provides, to our knowledge, the first evidence that human reaching may be governed not by static cost weights, but by a temporally structured cost landscape balancing effort, smoothness, and accuracy throughout the movement.


\section{Conclusion}
This work introduced a modified MO--IRL framework for learning time-varying cost functions that predict human reaching movements with substantially higher accuracy than classical fixed-weight approaches. The inferred cost structure consistently highlights the dominant role of joint acceleration, complemented by a smaller contribution of torque-change smoothness in the middle of the motion, reflecting a multi-criteria optimality principle that adapts throughout the movement. For the first time a single subject- and posture-independent cost function was identified and its ability to predict motion demonstrated with real experimental data. It suggests that reaching may be governed by a shared temporal organization rather than by fixed task-specific parameters.

A central methodological contribution of this study is the use of both joint positions and joint velocities during weight estimation. Incorporating velocity information reduces the ambiguity inherent in position-only formulations and improves the identifiability of dynamic cost terms. This points toward a broader conclusion: future IOC and IRL frameworks may benefit from exploiting the full state and input trajectories, including joint torques, external forces, and interaction-based cues, especially when dealing with inherently multimodal human motion.

An important next step is to assess whether the proposed cost function generalizes to more realistic reaching scenarios involving full 3D arm motion. Extending the framework to higher-dimensional models will allow testing whether the temporal structure uncovered here persists beyond planar tasks, and whether additional cost components emerge when multi-joint coordination become more complex.

From a robotics perspective, the inferred time-varying cost structure provides a principled foundation for designing bio-inspired controllers and generating human-like trajectories in humanoid and collaborative robots. A key advantage of MO--IRL is its ability to learn generalizable cost functions from only a small number of demonstrations. This greatly reduces the data-collection burden typically associated with human motion modeling and could enable the rapid construction of large-scale synthetic motion datasets. Such datasets are increasingly valuable for training robot imitation-learning policies, for benchmarking perception or prediction modules, and for providing diverse, human-consistent motion samples in simulation.

\bibliographystyle{Transactions-Bibliography/IEEEtran}
\bibliography{ref}

\end{document}